\documentclass[conference]{IEEEtran}

\usepackage{cite}
\usepackage{amsmath,amssymb,amsfonts}

\usepackage{algorithm,refcount}
\usepackage[noend]{algpseudocode}

\usepackage{graphicx}
\usepackage{textcomp}
\usepackage{xcolor}

\usepackage{hyperref}
\usepackage{subcaption}

\usepackage[bottom,flushmargin,hang,multiple]{footmisc}

\begin{document}

\title{Optimizing Grouped Convolutions on Edge Devices}

\author{\IEEEauthorblockN{Perry Gibson\IEEEauthorrefmark{1},
Jos\'e Cano\IEEEauthorrefmark{1},
Jack Turner\IEEEauthorrefmark{2},
Elliot J. Crowley\IEEEauthorrefmark{2},
Michael O'Boyle\IEEEauthorrefmark{2},
Amos Storkey\IEEEauthorrefmark{2}}
\IEEEauthorblockA{
        \begin{tabular}{cc}
            \IEEEauthorrefmark{1}{University of Glasgow, UK} 
            \IEEEauthorrefmark{2}{University of Edinburgh, UK}
        \end{tabular}
   }
}

\maketitle

\begin{abstract}

When deploying a deep neural network on constrained hardware, it is possible to replace the network's standard convolutions with grouped convolutions. This allows for substantial memory savings with minimal loss of accuracy. However, current implementations of grouped convolutions in modern deep learning frameworks are far from performing optimally in terms of speed. In this paper we propose \emph{Grouped Spatial Pack Convolutions} (GSPC), a new implementation of grouped convolutions that outperforms existing solutions. We implement GSPC in TVM, which provides state-of-the-art performance on edge devices. We analyze a set of networks utilizing different types of grouped convolutions and evaluate their performance in terms of inference time on several edge devices. We observe that our new implementation scales well with the number of groups and provides the best inference times in all settings, improving the existing implementations of grouped convolutions in TVM, PyTorch and TensorFlow Lite by 3.4$\times$, 8$\times$ and 4$\times$ on average respectively. Code is available at \texttt{https://github.com/gecLAB/tvm-GSPC/}

\vspace{1mm} 

\end{abstract}

\section{Introduction}

The deployment of deep neural networks onto mobile and embedded edge devices (e.g. IoT boards, smartphones, robots, drones, etc) has been relatively slow due to the resource constraints under which these devices operate. Big, memory-intensive neural networks typically perform poorly on edge devices. Because of this, there is a wealth of work concerned with compressing these networks. However, the predominant focus on the machine learning side of the problem, where the main performance metric considered is accuracy, has led to a proliferation of methods with promising compression results but non-trivial implications for hardware efficiency. That is, many neural architecture compression techniques may not work as expected at the system level where one of the main metrics considered is the inference time. Turner et al.~\cite{iiswc_2018} demonstrated that compression at the neural architecture level may have negative effects further down the \emph{Deep Learning Inference Stack}, depending on the choices of algorithmic transformation and the target hardware device. 

\begin{figure}[t]
 \centering
 \includegraphics[width=0.92\columnwidth]{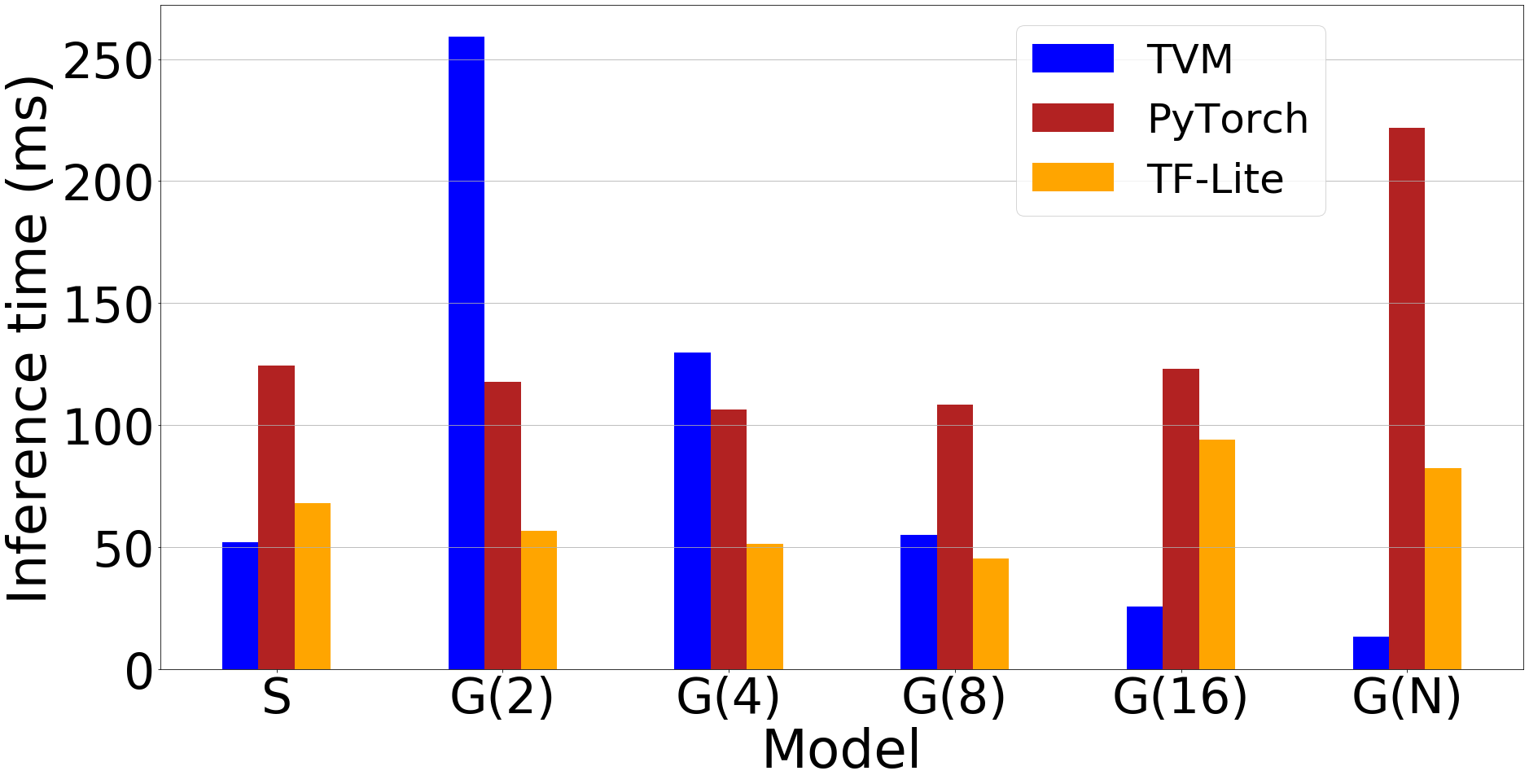}
 \caption{Inference time for WideResNet models using standard (S) and grouped convolutions ($G$) on the CPU of the Hikey 970 board for three common deep learning frameworks. No framework shows the expected behavior: i) faster execution than standard when using grouped convolutions, $G(g)$ where $g$ is the number of groups; ii) the time decreases as the number of groups increases.}
 \label{fig:motivation}
\end{figure}

Many compression techniques revolve around replacing the standard convolutions in a neural network with grouped convolutions~\cite{howard2017mobilenets,sandler2018MobileNetV2,moonshine,huang2018condensenet}. These allow for substantial savings in memory with minimal loss of accuracy, and are becoming increasingly prevalent. However, when evaluating the implementation of grouped convolutions present in current state-of-the-art deep learning frameworks such as PyTorch~\cite{paszke2019pytorch}, TensorFlow Lite~\cite{tflite} and TVM~\cite{tvm}, we observe that the measured inference times are far from the expected ones. Figure \ref{fig:motivation} shows an initial experiment where we run WideResNet models using standard (\emph{S}) and grouped convolutions (\emph{G}) on the CPU of the Hikey 970 board for the previous frameworks (note that we use~\emph{$G(g)$} to denote a grouped convolution using~\emph{g} groups). As we can see, none of the frameworks provides the theoretical expected behavior, that is: i) models using grouped convolutions should execute in less time than the initial model implementing standard convolutions (\emph{S}), since the overall number of Multiply-Accumulate (MAC) operations decreases when using groups (see Section~\ref{sec:modelcompression}); ii) as the number of groups increases (i.e. 2, 4, 8, etc), the number of MACs decreases and thus the execution time should also decrease. 

Motivated by Figure \ref{fig:motivation}, we propose a new implementation of grouped convolutions, that we call \emph{Grouped Spatial Pack Convolutions} (GSPC), which: i) outperforms all the previous implementations of grouped convolutions present in current deep learning frameworks; ii) brings us closer to the theoretical optimal performance level according to the reduction in the number of MAC operations. As we will see in Section~\ref{sec:exp}, although the behavior of our solution is consistent, there is still a small gap with respect to the theoretical expected times (we leave the investigation of this for future work). Note that since TVM is currently considered the state-of-the-art framework in terms of inference performance on edge devices, we integrate our GSPC implementation into the TVM source code. 

The contributions of this paper are as follows:

\begin{itemize}
    \item We propose a new algorithm for grouped convolutions, GSPC, and implement it in TVM.
    
    \item We evaluate the performance of GSPC using different network models on the CPU of several edge devices.
    
    \item We compare GSPC against implementations of grouped convolutions present in widely used deep learning frameworks, and we show that our solution outperforms them.
    
    \item We quantify the performance gap between the theoretically expected inference times and the measured ones. 

\end{itemize}

We briefly describe grouped convolutions in Section~\ref{sec:modelcompression}. In Section~\ref{sec:conv} we discuss our GSPC and the details of the specific implementation in TVM. Section~\ref{sec:exp} shows our experimental setup and a performance evaluation of several networks with grouped convolutions on three edge devices, discussing the time/accuracy trade off, analyzing different implementations on TVM and comparing GSPC with other existing implementations of grouped convolutions. In Section~\ref{sec:rw} we discuss previous related works. Finally, Section~\ref{sec:conclusion} concludes the paper and briefly discusses potential ideas for future work.

\begin{figure}[t]
 \centering
 \includegraphics[width=0.80\columnwidth]{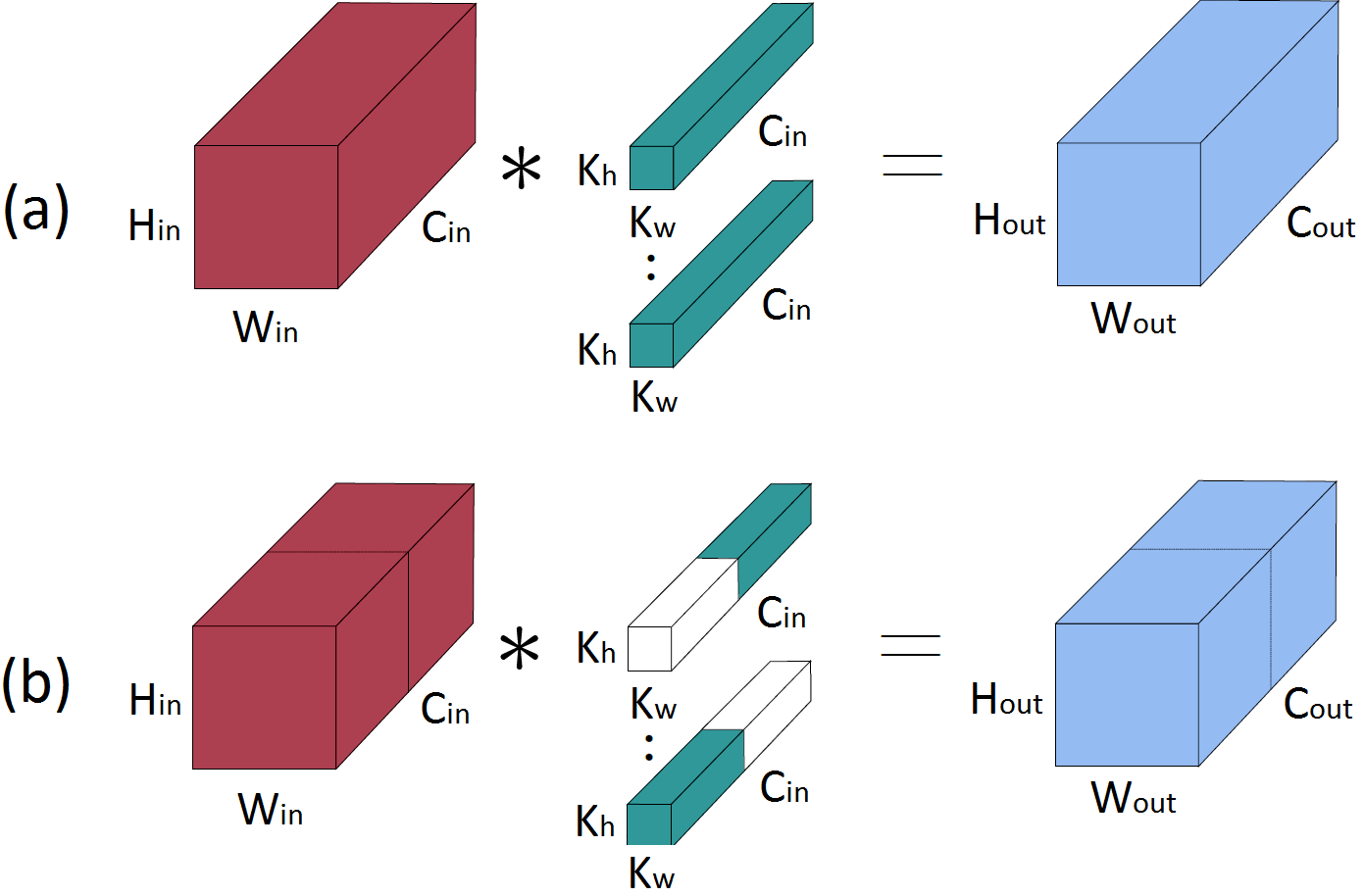}
 \caption{Standard vs. grouped convolutions: (a) In a standard convolution
~\emph{S}, each filter is convolved with all of the input's channels; (b) In a grouped convolution with two groups~\emph{$G(2)$}, half of the filters are applied to each half of the input for a $2 \times$ reduction in parameters used. More generally, a grouped convolution with $g$ groups uses $g \times$ fewer parameters.}
 \label{fig:grouped-conv}
\end{figure}

\section{Compression via Grouped Convolutions}
\label{sec:modelcompression}

Many works~\cite{howard2017mobilenets,sandler2018MobileNetV2,deeproots17,moonshine,huang2018condensenet,Turner2020BlockSwap} have demonstrated the efficacy of replacing standard convolutions $S$ with grouped convolutions for network compression. We denote these grouped convolutions as $G(g)$ where $g$ is the number of groups. This allows for a trade off between increased model compression against reduced accuracy as we increase $g$.

Consider a standard convolution as depicted in Figure~\ref{fig:grouped-conv}(a): its input consists of $C_{\mathrm{in}}$ channels. Each of $C_{\mathrm{out}}$ filters is convolved with {\bf all} of these input channels to produce a single channel filter output. These outputs are concatenated to give the $C_{\mathrm{out}}$ channel output of our convolution. Each filter uses $ C_{\mathrm{in}} \times K_{h} \times K_{w}$ parameters, where $K_{h} \times K_{w}$ is the kernel size of our filters. As we have $C_{\mathrm{out}}$ filters in total then the overall parameter cost is $C_{\mathrm{out}}\times C_{\mathrm{in}} \times K_h \times K_w$.

Now, consider instead the case where {\bf half} of our $C_{\mathrm{out}}$ filters are convolved with the first $C_{\mathrm{in}}/2$ channels of the input, and the other half of our filters are convolved with the second $C_{\mathrm{in}}/2$ channels of the input, as depicted in Figure~\ref{fig:grouped-conv}(b). Our filters now only use half as many parameters since each is now of size $ C_{\mathrm{in}}/2 \times K_h \times K_w$. This is a grouped convolution using two groups, and the total parameter cost is two times less than a standard convolution. For $g$ groups the parameter cost is reduced by a factor of $g$. A disadvantage of this grouping is that this prevents channels from mixing across groups; to counter this, practitioners typically follow grouped convolutions with a pointwise (kernel size 1) convolution~\cite{howard2017mobilenets}, which incurs an additional $C_{\mathrm{out}} \times C_{\mathrm{out}}$ parameter cost.

\begin{figure*}[t]
\centering
  \includegraphics[width=0.88\linewidth]{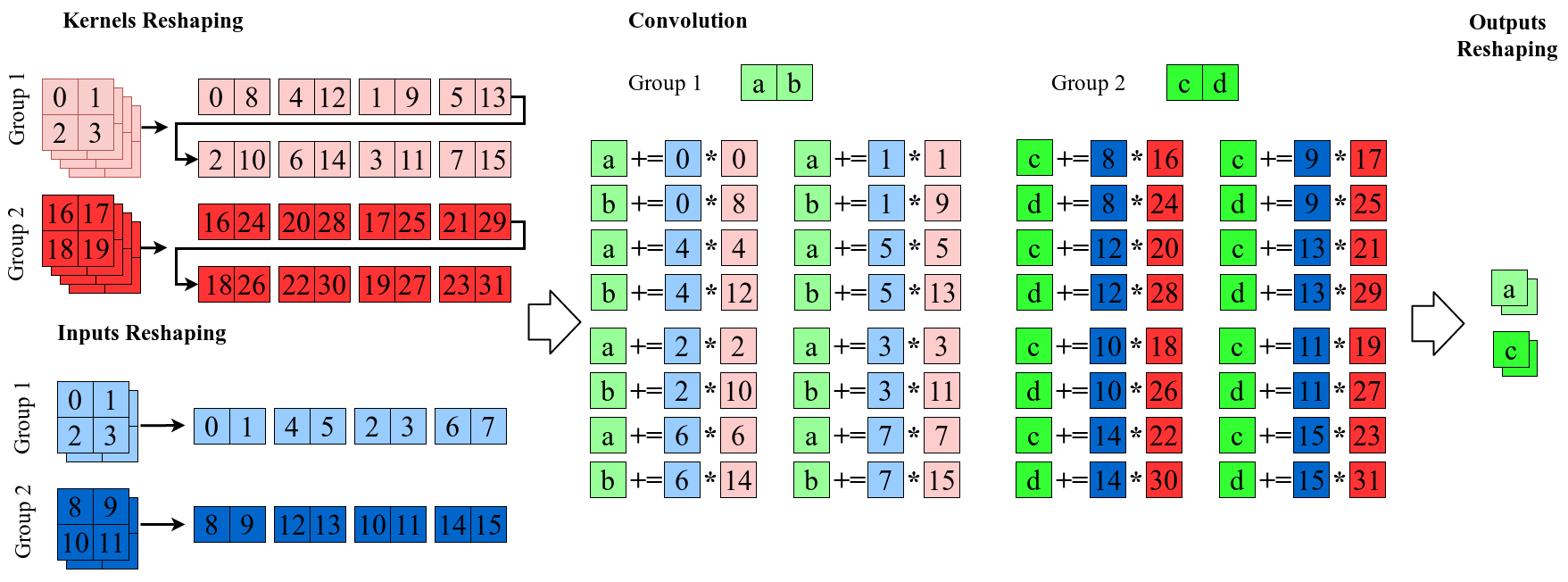}
  \caption{Overview of GSPC using a concrete example: two groups, four filters, four input channels, strides of one, and no input padding. Note that the input size is $1\times4\times2\times2$, the kernels size is of dimension $4\times2\times2\times2$ and numbers represent indices in the \texttt{NCHW} format.}
  \label{fig:gsp_example}
\end{figure*}

\section{Grouped Spatial Pack Convolutions}
\label{sec:conv}

\subsection{Motivation}

There are a number of popular algorithms that implement standard convolutional layers in neural networks, with a myriad of trade-offs.  For example, direct convolution is the simplest conceptually, requiring no data reshaping, and processes data using the ``sliding window'' that the convolution operation is often described as.  Other approaches like GEMM convolution reshape and sometimes expand inputs or weights (which can improve performance by improving the locality of data), however this increases memory footprint and the reshape stages may have a non-negligible contribution to the inference time.  For grouped convolutions, the algorithms available and their potential trade-offs have been explored less.  

In TVM, the algorithm used for grouped convolutions on CPUs is grouped direct convolution.  This is a variant of the standard direct convolution algorithm, and similarly has the advantage of requiring no additional memory footprint.  However, without any reshaping of data or weights, it suffers from poor data locality, and thus performance can be sub-optimal, especially for large layers since parameters may need to be reloaded from higher levels of cache.  For this reason, direct convolution is rarely used as the algorithmic primitive for standard convolutional layers.  However their effectiveness for grouped convolutions may have different considerations.

If we observe Figure \ref{fig:motivation}, we see that TVM's grouped direct convolution scales well as the number of groups increases, and outperforms both PyTorch and TensorFlow Lite for large values of $g$.  This scaling makes sense since the increased $g$ reduces the number of MACs for grouped convolutional layers in the model.  However, we observe that the time for $G(2)$, which reduces by $60$\% the number of MACs, is 4$\times$ slower than the $S$ model in TVM. Given the poor performance of $G(2)$, it is clear that an alternative approach to grouped convolutions is required to realize the potential performance improvements derived from the reduction in the number of MACs.

In this paper, we propose grouped spatial pack convolutions (GSPC), for the common \texttt{NCHW} data layout\noindent{\footnote{Input and output data are 4D arrays in row-major order, where $N$ is the batch size, $C$ is the number of channels, $H$ and $W$ are height and width.}}.  We modify and extend the spatial pack convolutions (SPC) algorithm described in \cite{zheng2018a}, which does not cover grouped convolutions.  Like SPC, GSPC reshapes data, kernels, and outputs to exploit data locality for the computation.  Our extension splits and computes data along an additional outer dimension for groups.  Since there is no data dependency between groups, this outer dimension can be leveraged to efficiently divide data between processing cores.  We implement the algorithm using TVM's tensor compute language, as it is portable across a wide variety of platforms, and can generate code which achieves state-of-the-art performance on many common deep learning benchmarks.  We favored implementing the GSPC algorithm in TVM, since the predictable scaling of its default grouped convolutions suggests that TVM's code generator would be more likely to give us reasonable scaling.

\subsection{General description}

At a high level, GSPC is comprised of four stages.  We expect a 4D input volume of size $NC_{\mathrm{in}}H_{\mathrm{in}}W_{\mathrm{in}}$, 4D kernels of size $C_{\mathrm{out}}C_{\mathrm{in}/g}K_h K_w$, and a 4D output volume of size $NC_{\mathrm{out}}H_{\mathrm{out}}W_{\mathrm{out}}$\footnote{These are standard layouts used for image data in convolutional neural networks.  The $_{\mathrm{in}}$ and $_{\mathrm{out}}$ subscripts indicate input and output dimensions respectively. The $C_{\mathrm{in}}/g$ dimension of the weights is one $g^{\textrm{th}}$ of the input dimensions (the number of parameters decreased as $g$ increases).}.  GSPC reshapes this data to improve locality.  The reshape has two values which represent tile size: $T_O$ and $T_I$, the former for tiling across output channels and the latter for tiling across input channels. Note that data in the same tile is related, and thus can enable further optimizations such as vectorization.  We define $\mathrm{\textit{KPG}}$ to be the number of kernels per group, and $\mathrm{\textit{CPG}}$ as the number of input channels per group. The four stages of the GSPC algorithm are: 

\begin{itemize}
    \item Reshape 4D kernels into a new 7D volume: $C_{\mathrm{out}}C_{\mathrm{in}/g}K_h K_w \rightarrow g\lfloor \frac{\mathrm{\textit{KPG}}}{T_O} \rfloor\lfloor \frac{\mathrm{\textit{CPG}}}{T_I} \rfloor K_h K_w T_I T_O$.
\vspace{1mm}
   
    \item Reshape 4D padded input data into a new 6D volume: $NC_{\mathrm{in}}H_{\mathrm{in}}W_{\mathrm{in}} \rightarrow gN\lfloor \frac{\mathrm{\textit{CPG}}}{T_I} \rfloor H_{\mathrm{in}} T_I W_{\mathrm{in}}$.
\vspace{1mm}

    \item Perform the grouped convolution using the 7D weights and 6D inputs, storing the output in a temporary 6D volume.  The computed 6D volume is of shape $gN \lfloor \frac{\mathrm{\textit{KPG}}}{T_O} \rfloor H_{\mathrm{out}} W_{\mathrm{out}} T_O$.
\vspace{1mm}
   
    \item Reshape the 6D output volume to the desired 4D output.
\end{itemize}

The kernel reshaping stage can be computed ahead of time and stored on disk in lieu of the default layout, since it does not depend on the input data.  By reordering our weights and inputs, we can improve the memory locality of our computation, which can reduce the cost of loads for elements being computed on.  Similarly, accumulating the convolution on a 6D intermediate array, and reshaping to 4D output is preferable to accumulating directly onto 4D as improved locality can improve cache behavior. The tile sizes are constrained by:

\vspace{-2mm}

\begin{equation} \label{eq:tile_constraints}
\begin{split}
0 &< T_O \leq \mathrm{\textit{KPG}} \\
0 &< T_I \leq \mathrm{\textit{CPG}}
\end{split}
\end{equation}

However, the ideal values for these tiles can vary. In the description of the stages of GSPC, observe that the inner dimension of the reshaped kernels and the intermediate outputs is $T_O$. Thus, the SIMD lane size for the target CPU can be a reasonable default for $T_O$, since data in the inner dimension is adjacent in memory and can thus be easily vectorized. 

Figure \ref{fig:gsp_example} illustrates the GSPC algorithm with an example.  We use tile sizes $T_O = T_I = 2$, as these are the maximum values allowed by the constraints (\ref{eq:tile_constraints}).  The initial data layout is shown on the left, with the channels split by group for clarity. The 6D and 7D volumes are shown flattened.  We observe how the input data and kernels are reshaped to improve data locality.  Even in the original data layout, data is divided between groups and the GSPC reshape stages maintain this division.  The MAC operations can be ordered to reduce the number of loads for each tile. In this example, each input value is used twice, thus computing these MACs together is a load-efficient approach.  The outputs reshaping stage is trivial in this case due to the small output size, and thus from a 1D memory perspective the reshape is the identity.  In the case of depthwise convolution, output reshaping is also the identity, which saves $N\times C_{\mathrm{out}} \times H_{\mathrm{out}} \times W_{\mathrm{out}}$ copy operations. 

Algorithm \ref{alg:gsp} describes the grouped spatial pack convolutions for the \texttt{NCHW} layout.  The stride hyperparameter is defined with $S_h$, $S_w$.  Note that the number of input and output channels should be divisible by the number of groups, so we can evenly split data between the groups.  The costs of the complex index arithmetic is reduced by TVM's ahead-of-time compilation for individual workloads. This means that expressions involving constants are simplified, which greatly reduces the number of computations at runtime.

The main loop of kernel reshaping (lines $2-4$) is nested with depth 7, with each loop representing a dimension of the reshaped kernel volume.  The same is true for inputs reshaping ($6-8$), with a nested loop depth of 6.  The main convolution loop ($10-18$) is over the 6 dimensions of the temporary output volume, with an additional 3 loops over $\mathrm{\textit{CPG}}$ and the kernel dimensions. For performance, these loops can be reordered.

\begin{algorithm}[t]
\small
  \caption{Grouped Spatial Pack Convolutions (GSPC)} 
  \label{alg:gemm-conv}
  \begin{algorithmic}
    \State $X:$ inputs of shape $NC_{\mathrm{in}}H_{\mathrm{in}}W_{\mathrm{in}}$
  \State $W:$ kernels of shape $C_{\mathrm{out}}C_{\mathrm{in}/g}K_h K_w$
  \State $T_I$, $T_O$: Tile sizes (constrained by equation \ref{eq:tile_constraints})
    \State $\mathrm{\textit{KPG}} \gets \frac{C_{\mathrm{out}}}{g}$, \quad $\mathrm{\textit{CPG}} \gets \frac{C_{\mathrm{in}}}{g}$
  \end{algorithmic}
  \textbf{Kernels Reshaping}
  \begin{algorithmic}[1]  
    \State Allocate $W'$ of dimension $g \lfloor \frac{\mathrm{\textit{KPG}}}{T_O} \rfloor \lfloor \frac{\mathrm{\textit{CPG}}}{T_I} \rfloor K_h K_w T_I T_O$
    \For{Dimensions of $W': j,k,c,h,w,ci,co$}
    \State $w \gets W[c \times T_O + co + j \times \mathrm{\textit{KPG}}][c \times T_I + ci][h][w]$
    \State $W'[j][k][c][h][w][ci][co] \gets w$ \label{alg:last-step1}
    \EndFor
  \end{algorithmic}

  \textbf{Inputs Reshaping}
  \begin{algorithmic}[1]
  \setcounterref{ALG@line}{alg:last-step1}
    \State Allocate $X'$ of dimension $g N \lfloor \frac{\mathrm{\textit{CPG}}}{T_I} \rfloor H_{i} T_I W_{i}$
    \For{Dimensions of $X': j,n,C,h,c,w$}
    \State $x \gets X[n][C \times T_I + c + \mathrm{\textit{CPG}} \times j][h][w]$
    \State $X'[j][n][C][h][c][w]  \gets x$
    \EndFor  \label{alg:last-step2}
  \end{algorithmic}

  \textbf{Perform convolution}
  \begin{algorithmic}[1]
    \setcounterref{ALG@line}{alg:last-step2}
    \State Allocate $Y'$ of dimension $g N \lfloor \frac{\mathrm{\textit{KPG}}}{T_O} \rfloor H_{o} W_{o} T_O$
    \For{Dimensions of $Y'$: $j,n,occ,oh,ow,ocv$}
    \State $y \gets 0$
    \For{$c = 0$ to $\mathrm{\textit{CPG}}$}
    \For{$kh = 0$ to $K_h$}
    \For{$kw = 0$ to $K_w$}
    \State $x \gets X'[j][n][\lfloor\frac{ic}{T_I}\rfloor][oh \times S_h+kh][c \bmod T_I][ow \times S_w+k_w]$
    \State $w \gets W'[j][occ][\lfloor\frac{ic}{T_I}\rfloor][kh][kw][c \bmod T_I][ocb]$
    \State $y \mathrel{+}= x \times w$
    \EndFor
    \EndFor
    \EndFor
    \State $Y'[j][n][occ][oh][ow][ocb] \gets y$

    \EndFor \label{alg:last-step3}
  \end{algorithmic}
  
  \textbf{Outputs Reshaping}
  \begin{algorithmic}[1]
      \setcounterref{ALG@line}{alg:last-step3}
    \State Allocate $Y$ of dimension $N C_{\mathrm{out}} H_{\mathrm{out}} W_{\mathrm{out}}$
    \For{Dimensions of $Y$: $n,c,h,w$}
    \State $y \gets Y'[\lfloor \frac{c}{\mathrm{\textit{KPG}}} \rfloor][n][\lfloor \frac{c}{T_O} \rfloor \bmod \lfloor \frac{\mathrm{\textit{KPG}}}{T_O} \rfloor][h][w][(c \bmod T_O) \bmod \mathrm{\textit{KPG}}]$
    \State $Y[n][c][h][w] \gets y$
    \EndFor
  \end{algorithmic}
  \label{alg:gsp}
\end{algorithm}

\subsection{Description of TVM compute and schedules}
\label{subsec:tvm-sche}

\begin{figure}[t]
 \centering
 \includegraphics[width=0.65\columnwidth]{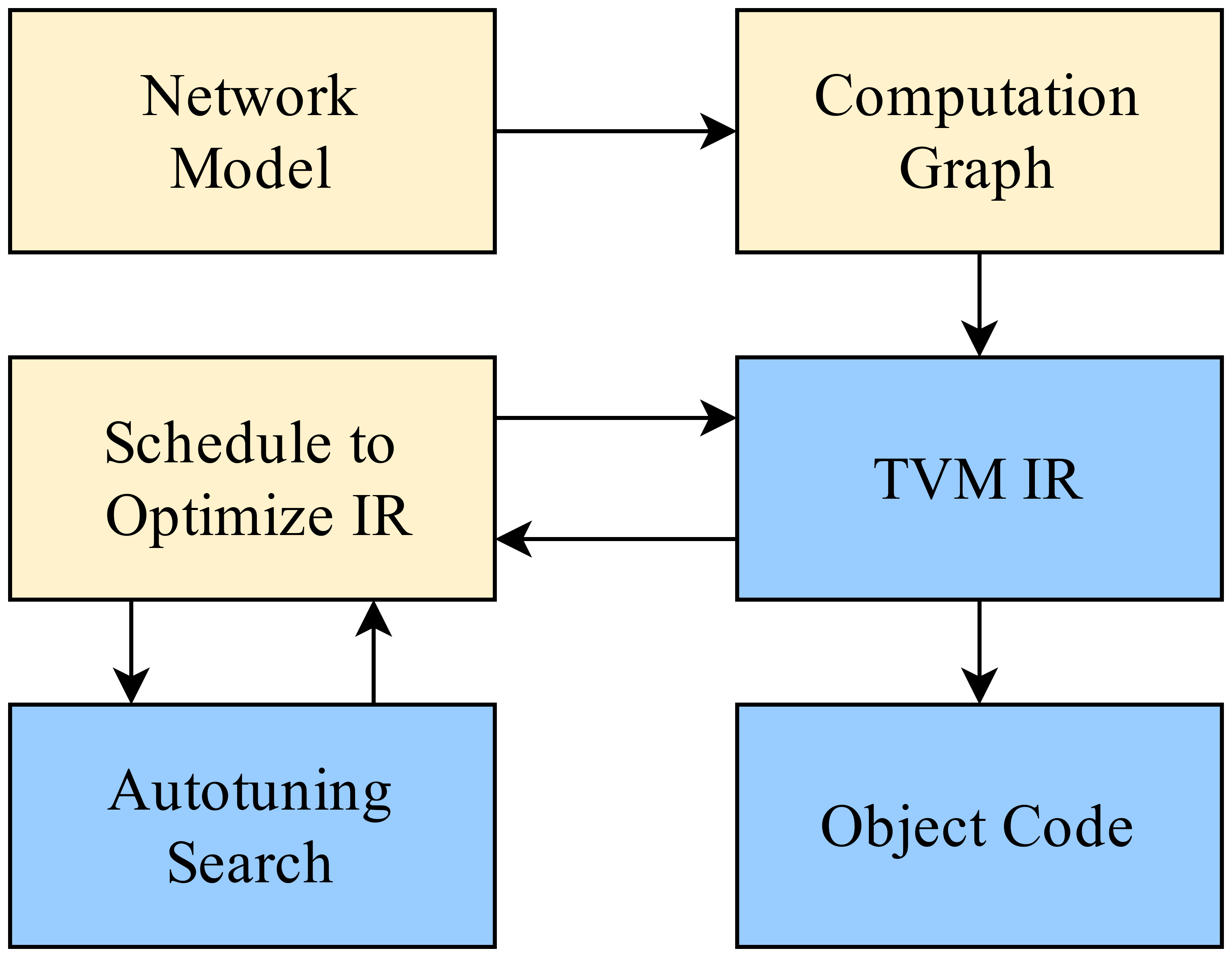}
 \caption{Overview of the relevant parts of TVM's stack. Light boxes are provided by us: including the benchmark models, and the definition for GSPC. Dark boxes are unaltered parts of the TVM compiler infrastructure.}
 \label{fig:tvm_sketch}
\end{figure}

We implement GSPC in TVM\footnote{The source code is available at https://github.com/gecLAB/tvm-GSPC/}, as it generates efficient code for tensor programs, provides the best time for the $S$ model, and scales well as $g$ increases, despite the poor performance of $G(2)$. Figure \ref{fig:tvm_sketch} gives an overview of the TVM stack. Expressing new algorithms in TVM has two main stages.

The first stage is describing the algorithm in TVM's compute language.  This is conveniently accessed via a Python API. These compute descriptions are used to generate TVM intermediate representation (IR) code which can be compiled to a variety of backends such as LLVM, OpenCL, and CUDA.  The algorithm implementation can be used as part of a computation graph generated from a neural network definition.

The second stage is manipulating the generated intermediate representation of the code to improve performance.  This uses TVM's schedule language, which is also accessible via a Python API.  A schedule is a set of transformations applied to the IR for a given algorithm, and can be specialized for a given platform (e.g. CUDA, x86, ARM CPU, etc).  This includes primitives for applying loop unrolling, loop fusion, loop fission, loop reordering, vectorization, and parallelization.

A well designed schedule is key in reaching near optimal performance.  In our experience, even small changes in the schedule have yielded large changes in the speed of the generated code, sometimes by an order of magnitude.  However, the potential for improvement from a schedule is constrained by properties of the underlying algorithm.  For example, the poor data locality of the standard direct convolution algorithm means that only modest improvements can be achieved.  Whereas in some algorithms, for example GSPC, even simple transformations can yield large performance improvements.  We reorder the nested loops of the convolution to make $T_O$ the innermost loop.  This can be leveraged for vectorization, hence why $T_O$ is set to the SIMD lane size of the target CPU by default if this is permitted by the constraint.

The schedule language includes primitives to expose various aspects of the schedule as tunable parameters that can be adjusted for different data shapes on a given platform to improve performance.  For example, exploring loop reordering may reduce inference time for some layer configurations but not others, or unrolling an inner loop may or may not give a performance boost depending on the number of iterations.  A strength of TVM is the autoTVM project \cite{chen2018c}, which allows exploration of this tuning space (autotuning) to improve inference time performance for target architectures.  For large models, the autotuning process is very time consuming, as the search space can be very large (e.g. individual models can take hours to tune, especially on constrained edge devices).  Some autotuning algorithms can also get stuck in suboptimal manifolds of the search space and require restarting.  More efficient autotuning, including the use of transfer learning to reuse knowledge about successful configurations for similar platforms or data shapes, is an area of active research. 

Tuning parameters exposed by our GSPC schedule include varying the tile sizes, and optionally unrolling the $K_w$ loop of the convolution stage.  There may be scope for additional improvements to the GSPC schedule, which could further reduce inference time. For example, a potential optimization could investigate the impact of interleaving portions of the reshaping and computation stages to reduce the footprint of the intermediate arrays by reusing a subset of their memory.

\begin{table*}[ht]
\caption{Hardware features of the devices used in the experiments.}
\begin{center}
\begin{tabular}{|c|c|c|c|c|c|}
\hline
\textbf{Device} & \textbf{CPU} & \textbf{L1 Cache (I+D)} & \textbf{L2 (+L3) Cache} & \textbf{RAM} & \textbf{Instruction Set} \\
\hline
Desktop & \hfill Intel i7-8700 (6 cores) @ 3.2 GHz & 192K + 192K & 1.5M (+12M) & 16GB DDR3 & x86 64-bit \\
\hline
HiKey 970 & Arm Cortex-A73 (4 cores) @ 2.4 GHz & 256K + 256K & 2M shared & 6GB LPDDR4 & ARMv8-A 64-bit \\
 & Arm Cortex-A53 (4 cores) @ 1.8 GHz & 128K + 128K & 1M shared & & \\
\hline
Raspberry Pi3B & Arm Cortex-A53 (4 cores) @ 1.2 GHz & 64K + 64K & 512K shared & 1GB LPDDR2 & ARMv8-A 64-bit \\
\hline
\end{tabular}
\label{table:platforms}
\end{center}
\end{table*}

\section{Evaluation}
\label{sec:exp}

\subsection{Experimental setup}

\subsubsection{Datasets and Networks} 

We consider two datasets widely adopted for image classification tasks, CIFAR-10~\cite{cifar} and ImageNet~\cite{ILSVRC15}, and we use the \texttt{float32} type to represent data values. We evaluate three deep neural network models, WideResNet-40-2 and ResNet-34 which are good representatives of residual network types, and MobileNetV2 which is a widely used model for edge devices. Some details of these models are the following:

\begin{itemize}
 \item WRN-40-2: we use a Wide Residual Network (WRN) \cite{zagoruyko2016wide} with 40 layers and width-multiplier 2 that requires 2.2 million parameters. We use the network as defined for CIFAR-10 classification.
 
 \item ResNet-34: we use a Residual Network with 34 layers \cite{he2016deep} that requires 21.8 million parameters. We use the ImageNet definition of the network.
 
 \item MobileNetV2: we use a Mobile Network \cite{sandler2018MobileNetV2} with 53 layers that requires 3.5 million parameters. We use its original ImageNet definition. 
\end{itemize}

We train networks for the previous three models where the standard convolutions are replaced with a grouped convolution followed by a pointwise standard convolution (as discussed in Section~\ref{sec:modelcompression}). We consider the following grouped convolutions: $G(g)~\forall~g \in \{2,4,8,16,N\}$, where $N$ is the number of input channels to each convolution. Note that the default architecture of MobileNetV2 uses $g=N$, so the original architecture is actually the MobileNetV2 $G(N)$ model. Also note that although pointwise convolutions incur a parameter cost, their inference time is negligible relative to grouped convolutions. This is because the operation is equivalent to a matrix multiplication over inputs and parameters, with a low number of MACs, and no data reshaping is required.

For WRN-40-2 and ResNet-34, networks were trained with attention transfer \cite{zagoruyko2016paying} on 1 and 4 NVIDIA TITAN X GPUs for 200 and 100 epochs respectively using Stochastic Gradient Descent (SGD) with momentum 0.9 to minimize cross-entropy loss, learning rate of 0.1 and weight decay 0.0005 (WRN-40-2) and 0.0001 (ResNet-34). For MobileNetV2, we trained networks with varying values of $g$ on 1 NVIDIA TITAN RTX GPU for 150 epochs using SGD with momentum 0.9, learning rate of 0.05 and weight decay 0.0004.

\subsubsection{Hardware platforms} 

The platforms used in this work are listed in Table~\ref{table:platforms}. There are two edge boards (Hikey 970, Raspberry Pi3B) that include both CPU and GPU, however in this work we focus on CPU evaluation, and we leave GPU investigation for future work. We also analyze a standard desktop CPU. Therefore, we evaluate Arm and Intel processors that implement two different instruction set architectures with frequencies ranging from 1.2GHz to 3.2GHz. Note that the CPU of the Hikey board implements the \emph{big.LITTLE}~\cite{biglittle} architecture (4 \emph{big} cores + 4 \emph{LITTLE} cores), but in this work we only use the \emph{big} cores. Finally, the memory hierarchy varies significantly across platforms. All these features give us a diverse set of configurations for evaluation.

\subsection{Speed vs Accuracy analysis}

\begin{table} [!tp]
\caption{Inference time in \textit{ms} for WRN-40-2 models with standard ($S$) and grouped ($G$) convolutions when running on the platforms in Table~\ref{table:platforms}.}
\begin{center}
\begin{tabular}{|c|ccc|ccc|}
\hline
\textbf{Model} & \textbf{Params} & \textbf{MACs} & \textbf{Top1} & \textbf{Desktop} & \textbf{HiKey} & \textbf{RPi3} \\
\hline
S & 2242.26K & 328.30M & 4.79 & 8.23 & 65 & 811 \\
\hline
$G(2)$ & 1357.68K & 198.15M & 4.87 & 9.20 & 51 & 530 \\
$G(4)$ & 813.36K & 118.52M & 5.00 & 5.84 & 34 & 307 \\
$G(8)$ & 541.20K & 78.71M & 5.05 & 4.65 & 24 & 199 \\
$G(16)$ & 405.12K & 58.80M & 5.13 & 4.51 & 20 & 158 \\
$G(N)$ & 292.22K & 44.83M & 6.57 & 2.14 & 16 & 122 \\
\hline
\end{tabular}
\label{table:wrn-1th}
\end{center}

\bigskip

\caption{Inference time in \textit{ms} for ResNet-34 models with standard ($S$) and grouped ($G$) convolutions when running on the platforms in Table~\ref{table:platforms}.}
\begin{center}
\begin{tabular}{|c|ccc|ccc|}
\hline
\textbf{Model} & \textbf{Params} & \textbf{MACs} & \textbf{Top1} & \textbf{Desktop} & \textbf{HiKey} & \textbf{RPi3} \\
\hline
S & 21.79M & 3.67G & 26.73 & 107 & 1096 & 7466 \\
\hline
$G(2)$ & 13.22M & 2.25G & 26.13 & 99 & 636 & 5700 \\
$G(4)$ & 8.14M & 1.39G & 26.58 & 62 & 426 & 3334 \\
$G(8)$ & 5.60M & 0.97G & 27.24 & 41 & 304 & 2344 \\
$G(16)$ & 4.34M & 0.75G & 27.99 & 34 & 259 & 1749 \\
$G(N)$ & 3.13M & 0.56G & 30.16 & 23 & 204 & 1285 \\
\hline
\end{tabular}
\label{table:resnet-1th}
\end{center}

\bigskip

\caption{Inference time in \textit{ms} for MobileNetV2 models with standard ($S$) and grouped ($G$) convolutions when running on the platforms in Table~\ref{table:platforms}.}
\begin{center}
\begin{tabular}{|c|ccc|ccc|}
\hline
\textbf{Model} & \textbf{Params} & \textbf{MACs} & \textbf{Top1} & \textbf{Desktop} & \textbf{HiKey} & \textbf{RPi3} \\
\hline
S & 44.05M & 5.56G & 26.03 & 166 & 1207 & 13770 \\
\hline
$G(2)$ & 23.75M & 2.92G & 25.90 & 135 & 776 & 7603 \\
$G(4)$ & 13.59M & 1.60G & 26.34 & 75 & 733 & 4608 \\
$G(8)$ & 8.52M & 0.95G & 26.84 & 47 & 495 & 2625 \\
$G(16)$ & 5.98M & 0.62G & 27.06 & 37 & 429 & 1808 \\
$G(N)$ & 3.50M & 0.31G & 28.20 & 15 & 134 & 812 \\
\hline
\end{tabular}
\label{table:mobilenet-1th}
\end{center}
\end{table}

\begin{figure*} [t]
\centering
  \begin{subfigure}[b]{0.32\linewidth}
    \centering
    \includegraphics[width=0.99\linewidth]{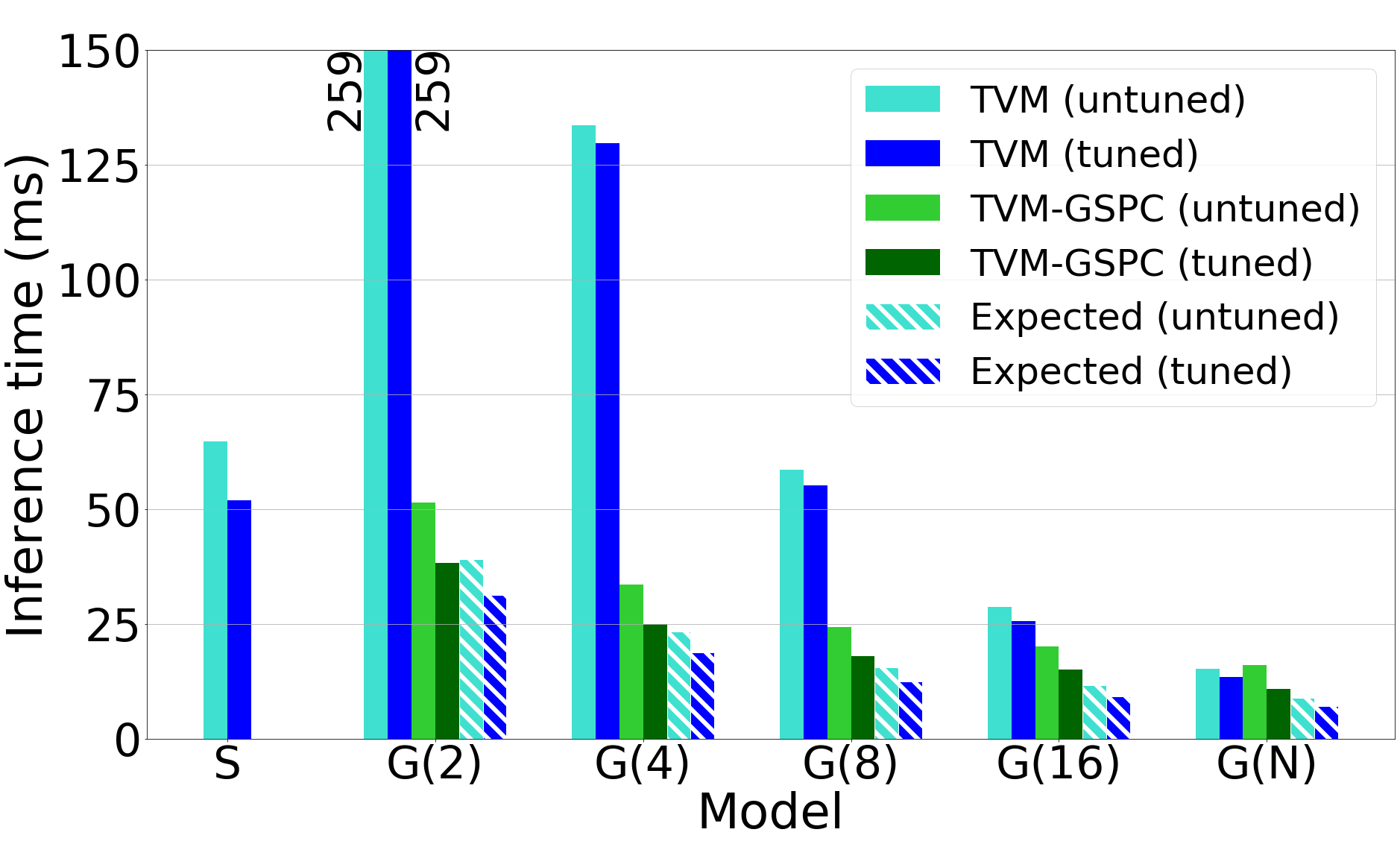}
    \caption{WRN-40-2}
    \label{fig:tvm_a}
  \end{subfigure}
  \begin{subfigure}[b]{0.32\linewidth}
    \centering
    \includegraphics[width=0.99\linewidth]{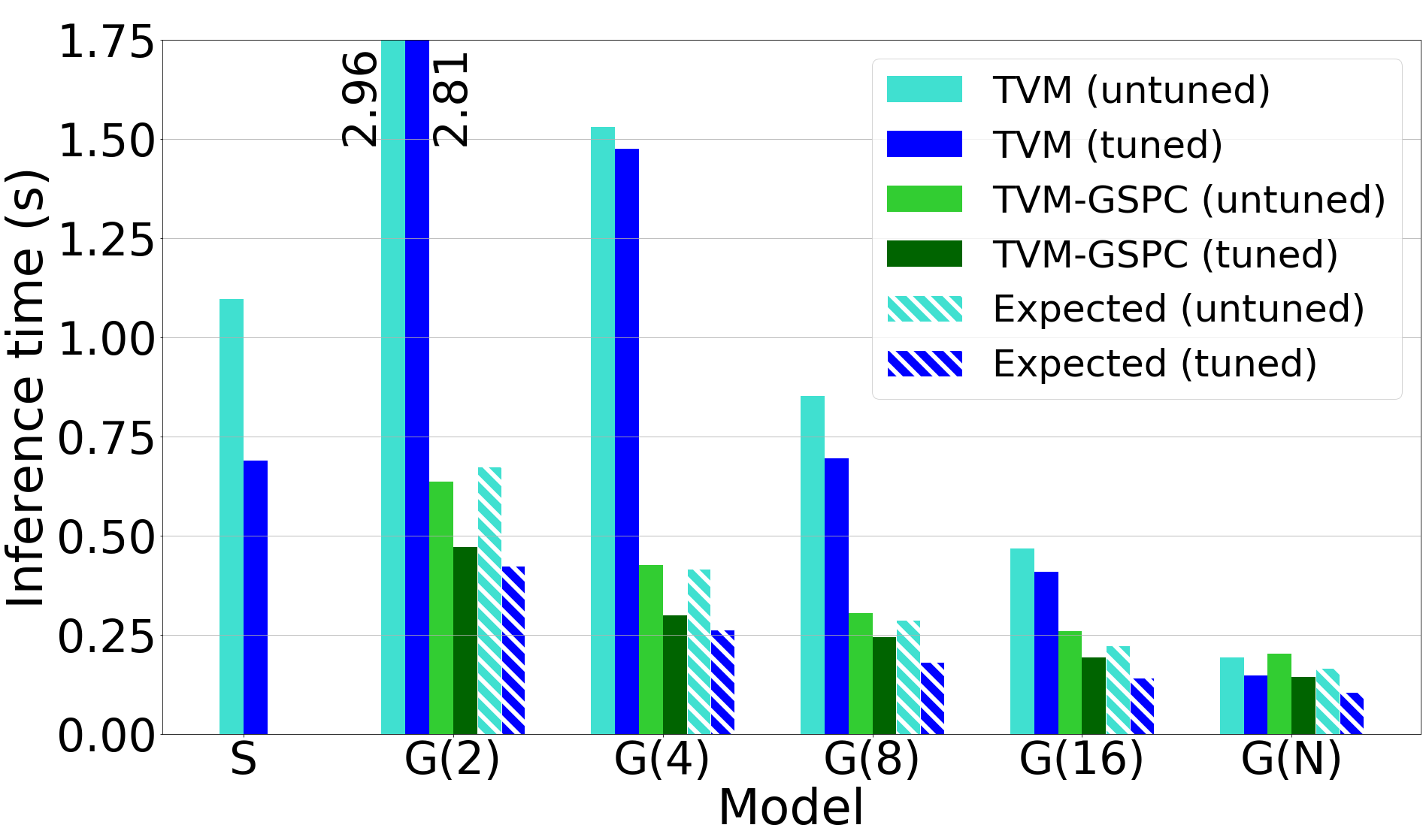}
    \caption{ResNet-34}
    \label{fig:tvm_b}
  \end{subfigure}
  \begin{subfigure}[b]{0.32\linewidth}
    \centering
    \includegraphics[width=0.99\linewidth]{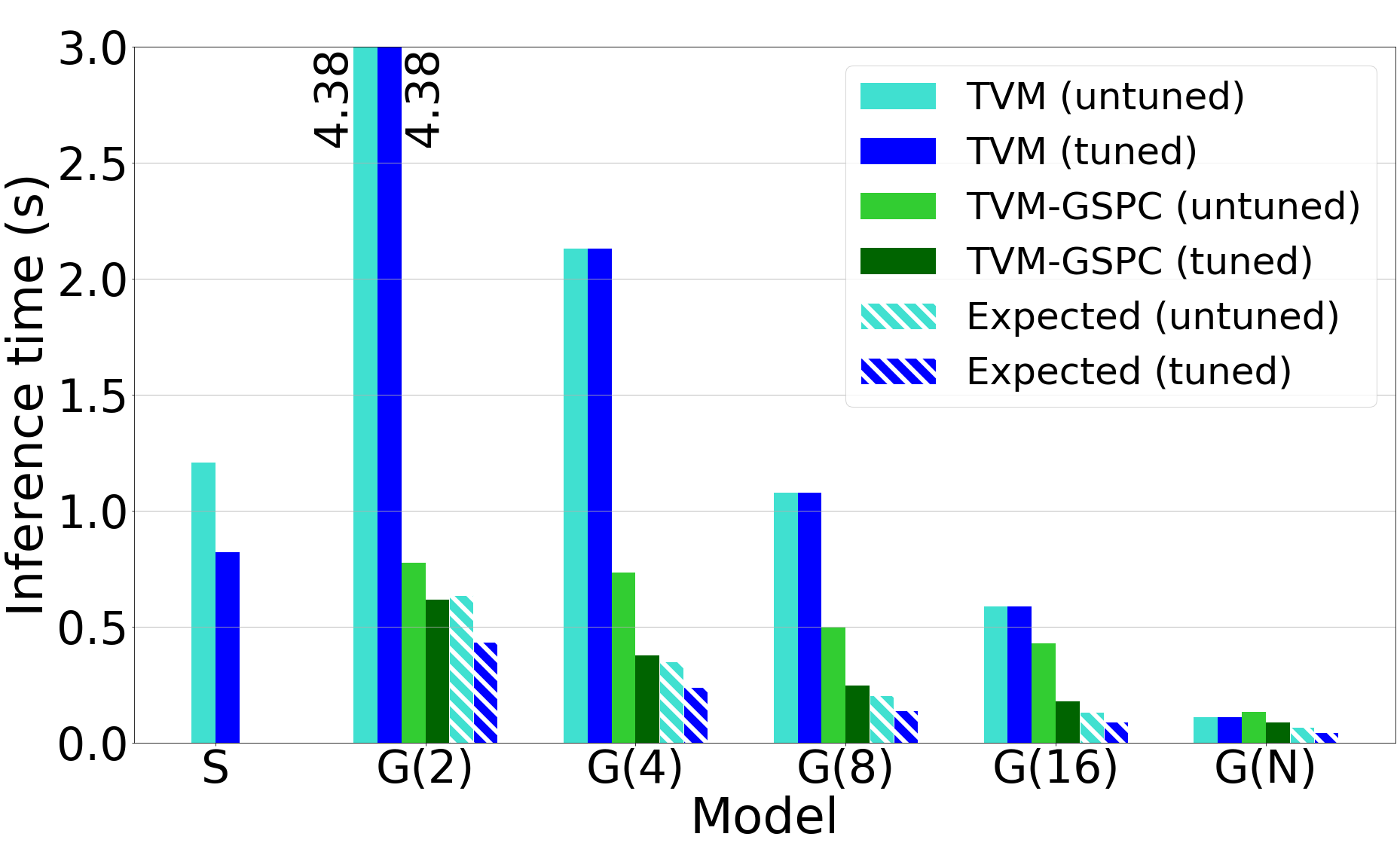}
    \caption{MobileNetV2}
    \label{fig:tvm_c}
  \end{subfigure}
\caption{Inference time in \textit{ms} for network models with standard ($S$) and grouped ($G$) convolutions when running on the CPU of the Hikey 970 board. We compare the measured and expected times of our GSPC and the default TVM implementation for both tuned and untuned versions of the code.}
\label{fig:TVM}
\end{figure*}

Tables \ref{table:wrn-1th}, \ref{table:resnet-1th} and \ref{table:mobilenet-1th} show the inference time in milliseconds for all the network models considered using standard ($S$) and grouped ($G$) convolutions for WRN-40-2, ResNet-34 and MobileNetV2 respectively when running on the three platforms under study\footnote{Note that all times are for 1 thread execution, since we verified that threads affect quite differently the performance of each platform, thus not providing a completely fair comparison. We leave the threads analysis for future work.} using our GSPC implementation in TVM. The tables also show the total parameter cost, the number of MACs, and the Top1 error of each network model. The number of MACs is obtained with the following formula:

\begin{equation}
\small
MACs = \frac{N \times C_{\mathrm{in}} \times C_{\mathrm{out}} \times K_h \times K_w \times H_{\mathrm{out}} \times W_{\mathrm{out}}}{g}
\end{equation}

where $N$ is the batch size ($N=1$ for all experiments), $C_{\mathrm{in}}$ is the number of input channels, $C_{\mathrm{out}}$ is the number of output channels, $H_{\mathrm{out}}$ and $W_{\mathrm{out}}$ are the height and width of the layer's output respectively, $K_h \times K_w$ is the kernel size of each convolution, and $g$ is the number of groups.

As we can see in the tables, the reduction in the number of parameters, and thus the number of MACs, derived from using grouped convolutions provides between \texttildelow$4$-$17\times$ of speedup in the inference time across platforms and networks, the Raspberry Pi device and the MobileNetV2 network being the combination that provides the highest improvements. We also observe that on the desktop the inference time for $G(2)$ is not reduced with respect to the corresponding $S$ model as on the other two platforms, it even increases for WRN-40-2. However, the time decreases for every subsequent $G$ model. This observation suggest that the schedule is not properly optimized for the Intel x86-64 architecture of the desktop. In TVM, the schedules can be optimized for a given hardware architecture and the default $S$ model is taking advantage of this, as we checked that it has schedules for both Intel and Arm architectures. However, we optimized the schedule of our GSPC code primarily for the Hikey platform, as we performed most of our experiments on it. Optimizing GSPC for the Intel architecture should provide better times for the $G$ models, but we leave this optimization for future work.

Related to the accuracy of the models, we see in Tables \ref{table:wrn-1th}, \ref{table:resnet-1th} and \ref{table:mobilenet-1th} that the increase in Top1 error can vary from almost $2$\% for WRN-40-2 to $3.5$\% for ResNet-34 when we compare the $S$ and $G$ models. We also see that the overall error is much higher for the models using the ImageNet dataset (\texttildelow$30$\% vs \texttildelow$7$\%), since a 1000-way classification is harder than a 10-way one. Therefore, these results provide different options to the user for selecting a model based on the time/accuracy trade off. The best solution for a given application will depend on its specific requirements and the hardware platforms available. For example, if the target platform is very constrained like the Raspberry Pi, it could be better to sacrifice some accuracy in favor of speeding up the inference time. However, for a more powerful platform like the desktop it can be better to maximize accuracy, as the all times are below $166$\textit{ms}.

\subsection{TVM analysis}
\label{subsec:tvm-analysis}

Figure \ref{fig:TVM} shows the \emph{Measured} versus the \emph{Expected\footnote{Computed from the inference time of the $S$ model on a given platform by obtaining the time of a single MAC operation and then extrapolating.}} inference time for all the models considered for the three networks under study when running on the Hikey 970 platform. We compare GSPC with the default implementation of grouped convolutions in TVM, and we consider the tuned (i.e. we use the autoTVM \cite{chen2018c} tool mentioned in Section~\ref{subsec:tvm-sche}) and untuned versions of the code in both cases. Note that the times in Tables \ref{table:wrn-1th}, \ref{table:resnet-1th} and \ref{table:mobilenet-1th} correspond to GSPC untuned times. The reason is that the tuning process is very time consuming and error prone, especially in constrained devices. Based on our tuning experience on the Hikey board, we estimate that getting the required tuned times on the Raspberry Pi board using the same number of tuning iterations would take weeks. Our key observations in Figure \ref{fig:TVM} are as follows: 

\begin{figure*}[t]
\centering
  \begin{subfigure}[b]{0.32\linewidth}
    \centering
    \includegraphics[width=0.99\linewidth]{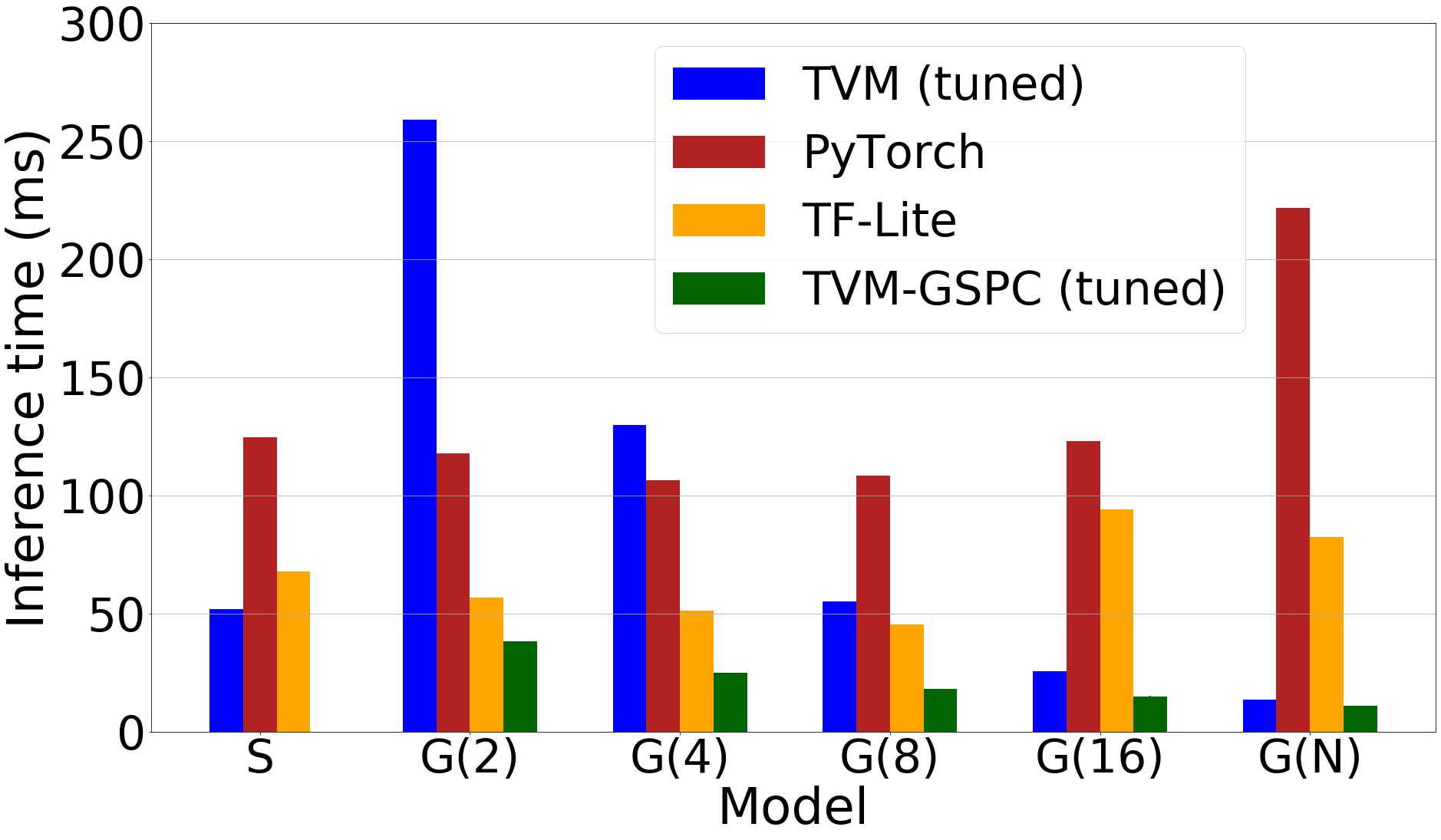}
    \caption{WRN-40-2}
  \end{subfigure}
  \begin{subfigure}[b]{0.32\linewidth}
    \centering
    \includegraphics[width=0.99\linewidth]{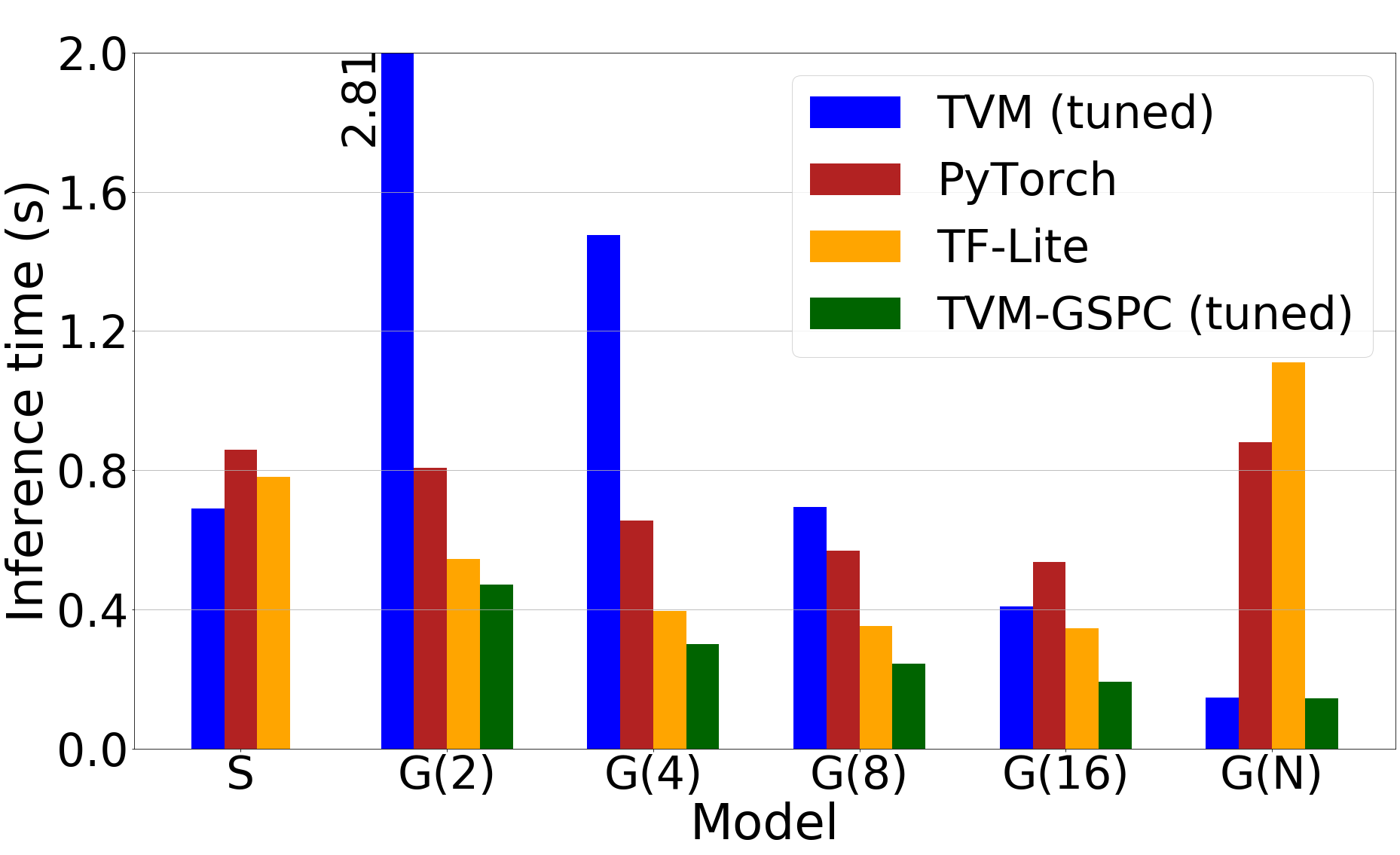}
    \caption{ResNet-34}
  \end{subfigure}
  \begin{subfigure}[b]{0.32\linewidth}
    \centering
    \includegraphics[width=0.99\linewidth]{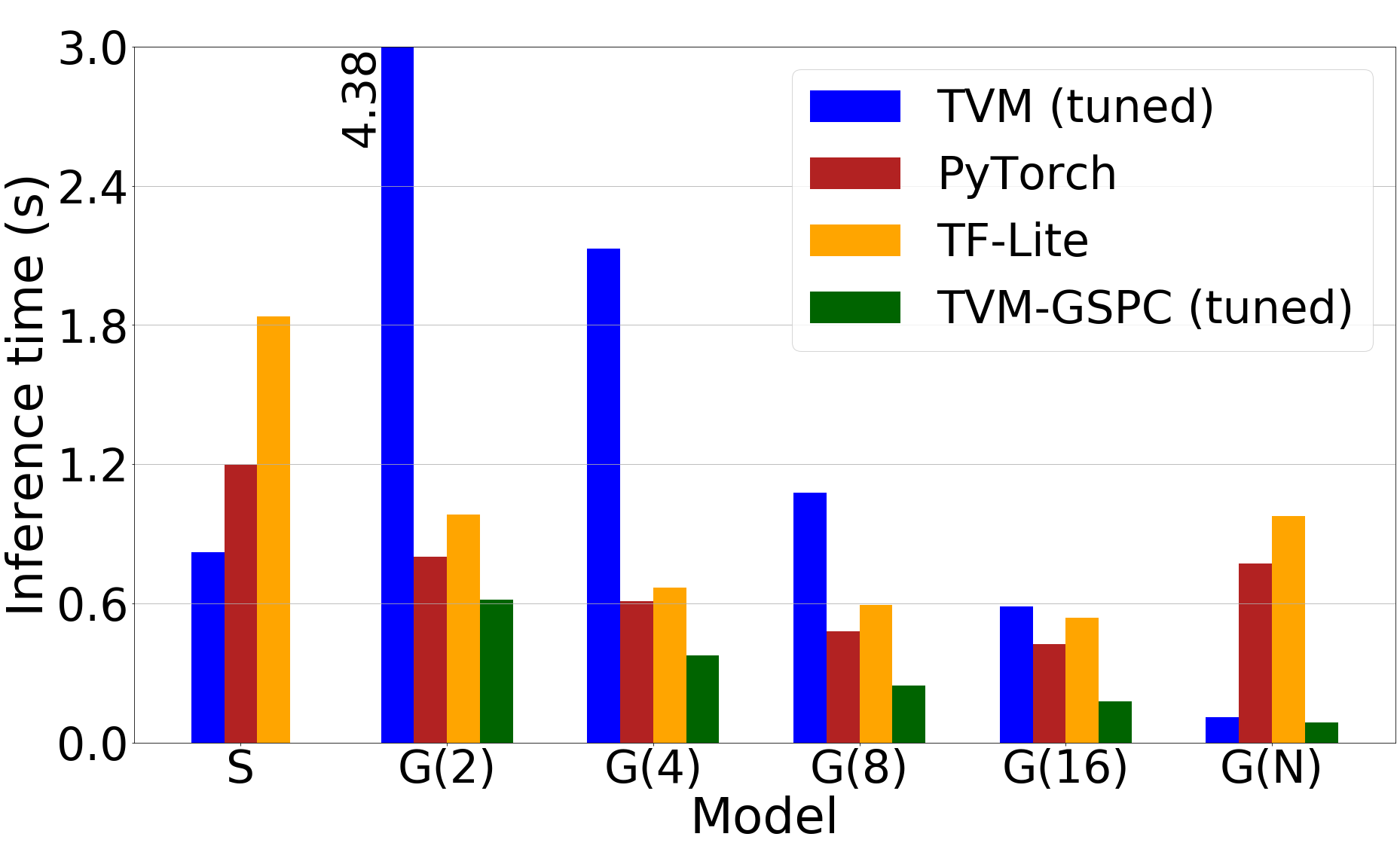}
    \caption{MobileNetV2}
  \end{subfigure}
\caption{Inference time in \textit{ms} for network models with standard ($S$) and grouped ($G$) convolutions when running on the CPU of the Hikey 970 board. We compare the tuned version of GSPC and default TVM against PyTorch and TensorFlow Lite.}
\label{fig:frameworks}
\end{figure*}

\begin{itemize}
    \item GSPC improves the times of the default TVM implementation of the $G(2)$-$G(16)$ models for the three networks for both tuned and untuned versions of the code. However, for $G(N)$ the default TVM implementation is slightly better than GSPC (\texttildelow$5$-$22$\% across networks) for the untuned version. Initially, we thought that the most likely reason for that could be the overhead created by the reshaping stages of GSPC, which for $G(N)$ are maximized relative to the computation time. However, we optimized these reshaping stages for our $G(N)$ models and we obtained the current differences. We leave for future work to investigate this problem further.
   
    \item When we consider the tuned versions of $G(N)$, GSPC provides better times than the default TVM implementation (\texttildelow$3$-$34$\% across networks). However, note that for MobileNetV2 the tuned times that we obtained for the default TVM implementation were worse than the untuned ones for all $G$ models. For this reason, in Figure \ref{fig:tvm_c} the tuned times are the same as the untuned ones. We could not find an explanation for this strange result, but we double-checked it running the tuner several times. 
   
    \item There are differences between the expected and measured times for both tuned and untuned versions across all $G$ models and networks. This performance gap is \texttildelow$7$-$99$\% for untuned and \texttildelow$27$-$72$\% for tuned versions respectively. Note that the expected times are theoretical estimations based on the structure of the code for the standard convolution, which should not be considered as true optimal values. In some cases, it can be possible to outperform the expected time (see $G(2)$ in Figure \ref{fig:tvm_b}), for reasons such as more data fitting in a lower level of cache.
\end{itemize}

Overall, our GSPC implementation is on average $3.4\times$ faster than the default TVM code for all the tuned/untuned $G$ models.

\subsection{Frameworks comparison}

Figure \ref{fig:frameworks} shows the inference time of GSPC and other implementations of grouped convolutions in current deep learning frameworks for all the $G$ models of the three networks under study when running on the CPU of the Hikey board. The figure also shows the times for the $S$ models. We consider the tuned version of both GSPC and default TVM. The other frameworks analyzed are PyTorch \cite{paszke2019pytorch} and TensorFlow Lite \cite{tflite}. 

As we can see, GSPC provides the best results for all the $G$ models of the three networks, clearly outperforming the default TVM and the other two frameworks, up to $8\times$ and $4\times$ better than PyTorch and TensorFlow Lite respectively. To the best of our knowledge, GSPC is the most efficient implementation, in terms of inference time, of grouped convolutions available. We also observe that TensorFlow Lite performs much better than PyTorch for all the $G$ models of WRN-40-2 and for the $G(2)$-$G(16)$ models of ResNet-34, whereas PyTorch is better for $G(N)$ of ResNet-34 and all the $G$ models of MobileNetV2. However, none of these frameworks scales as expected for the $G$ models according to the number of MAC operations.

\section{Related Work}
\label{sec:rw}

It is well-established that many modern neural networks are over-parameterized for inference~\cite{han2015learning}. A focus on achieving state-of-the-art results has led to bloated network architectures with diminishing returns when new parameters are added~\cite{huang2019gpipe}. Given that most of the energy consumption and execution time in such networks is dedicated to convolution~\cite{lai2018not}, it is desirable to reduce convolutional over-parameterization for use in resource-constrained settings. 
One popular method to exploit parameter redundancy is to split standard convolutions into {\it groups} along the channel dimension. These  {\it grouped convolutions} first appeared in AlexNet~\cite{alexnet} due to GPU memory constraints, and have since featured prominently in the network compression literature~\cite{howard2017mobilenets,deeproots17,moonshine,huang2018condensenet,sandler2018MobileNetV2}. 

As the number of groups is increased, the parameter cost of a grouped convolution decreases at the expense of representational capacity. In the extreme case where there are as many groups as there are convolutional channels we obtain {\it depthwise convolutions}. In MobileNets~\cite{howard2017mobilenets} for example, the authors replace standard convolutions with pairs of depthwise convolutions and pointwise ($1\times 1$) convolutions, the latter of which allows for channel mixing to restore capacity. The technique is known as depthwise separable convolutions \cite{sifre2014}. Similarly, in \cite{moonshine} the authors take the standard block used in ResNets~\cite{he2016deep} consisting of two standard convolutions, and replace each convolution with a grouped and pointwise pair. The number of groups can be varied to trade off accuracy against parameter total.

However, leveraging parameter reductions into better hardware performance remains difficult. Many frameworks transform convolution into matrix-matrix multiplication in order to exploit pre-existing, highly optimized subroutines~\cite{chetlur2014cudnn}, which may not be composable with any dimensionality perturbations caused by cheapening. Only recently has attention turned to direct convolution itself~\cite{georganas2018anatomy}, and though code generation frameworks promise performance portability for custom convolutions~\cite{vasilache2018tensor,tvm,venkat2019swirl,baghdadi2019}, they have been shown to lack the generality required to adopt such radical neural architecture changes to a vast hardware landscape~\cite{Barham:2019:MLS:3317550.3321441}.

Common approaches to convolution include direct convolution, Winograd convolution \cite{lavin2016}, and GEMM convolution. Winograd convolution involves mapping data into Fourier space to allow multiplications to become additions.  GEMM convolution is especially popular on GPUs, where the input data is expanded and reshaped using a method known as \textit{im2col}, so that convolution can be computed as a well optimized matrix multiplication with libraries such as OpenBLAS \cite{openblas} and ATLAS \cite{atlas_siam}. TVM's default approach to standard convolution on the CPU is an algorithm known as spatial packed convolution (SPC), described in \cite{zheng2018a}. Tradeoffs in performance can vary across platforms and architecture.

\section{Conclusion}
\label{sec:conclusion}

In this paper we have proposed Grouped Spatial Pack Convolutions (GSPC) as a new and more efficient implementation of grouped convolutions. We have implemented GSPC in TVM, which provides state-of-the-art performance on edge devices. We have evaluated several network models implementing grouped convolutions for two datasets on three edge devices with different hardware architectures. We have also compared our implementation against existing solutions in current deep learning frameworks, outperforming them in all settings. Finally, we observed that even though models based on grouped convolutions significantly improve the performance of the initial standard model, the expected inference time, based on the number of MAC operations, does not translate into measured performance. We leave to further study this performance gap for future work. We also leave for future work to analyze the performance of GSPC on the \emph{big.LITTLE} \cite{biglittle} architecture and embedded GPUs (e.g. Arm Mali) present in current edge devices. Additionally, translation of other approaches to standard convolution into the group convolution domain, e.g. Winograd, and investigation of their performance trade-offs across different benchmarks and devices. Finally, considering power \cite{chen2018b} is also an area for future research.

\section*{Acknowledgment}

This project received funding from the European Union's Horizon 2020 research and innovation programme under grant agreement No. 732204 (Bonseyes). This work was supported by the Swiss State Secretariat for Education, Research and Innovation (SERI) under contract number 16.0159. The opinions expressed and arguments employed herein do not necessarily reflect the official views of these funding bodies.

\bibliography{bib}

\bibliographystyle{IEEEtran}
\end{document}